\documentclass[10pt, a4paper]{scrartcl}

\usepackage{latexsym}
\usepackage{amssymb}
\usepackage[nonamelimits]{amsmath}
\usepackage[authoryear, round]{natbib}
\usepackage{url}

\usepackage{color}
\definecolor{myred}{rgb}{0.5,0,0}
\definecolor{myblue}{rgb}{0,0,0.75}
\definecolor{mygreen}{rgb}{0,0.5,0}

\usepackage{ifpdf}

\ifpdf
   \usepackage[pdftex]{graphicx}
   \usepackage[pdftex,
               pdftitle={},
               pdfsubject={},
               pdfkeywords={},
               pdfauthor={Dirk Tasche},
               pdfstartview=FitR,
               breaklinks=true,
               colorlinks=true,
               citecolor=myred,
               linkcolor=myblue,
               urlcolor=mygreen]{hyperref}
\else
   \usepackage[dvips]{graphicx}
   \usepackage{hyperref}
   \usepackage{rotating}
\fi

\newtheorem{theorem}{Theorem}[section]
\newtheorem{lemma}[theorem]{Lemma}
\newtheorem{remark}[theorem]{Remark}

\newtheorem{proposition}[theorem]{Proposition}

\newtheorem{corollary}[theorem]{Corollary}
\newtheorem{assumption}[theorem]{Assumption}

\newcommand{\qed}{$\Box$}

\numberwithin{equation}{section}

\title{Does quantification without adjustments work?}

\author{%
Dirk Tasche\thanks{E-mail: dirk.tasche@gmx.net\newline
The author currently works at the Swiss Financial Market Supervisory Authority (FINMA).
Most of the research for this paper was done while he was an employee of the Prudential Regulation Authority (a
directorate of the Bank of England) and on secondment to the Bank's research hub. 
The opinions expressed in this paper are those of the author 
and do not necessarily reflect views of FINMA or the Bank of England.}}

\date{August 12, 2016}

\begin{document}

\maketitle

\begin{abstract}
Classification is the task of predicting the class labels of objects based on the observation of their
features. In contrast, quantification has been defined as the task of determining the prevalences
of the different sorts of class labels in a target dataset. The simplest approach to quantification is 
Classify \& Count where a classifier is optimised for classification on a training set and 
applied to the target dataset for the prediction of class labels. In the case of binary quantification, the number of
predicted positive labels is then used as an estimate of the prevalence of the positive class in the target
dataset. Since the performance of Classify \& Count for quantification is known to be inferior
its results typically are subject to adjustments. However, some researchers recently have suggested
that Classify \& Count might actually work without adjustments if it is based on a classifier that
was specifically trained for quantification. We discuss the theoretical foundation for this claim
and explore its potential and limitations with a numerical example based on the binormal model with
equal variances. In order to identify an optimal quantifier in the binormal setting, we introduce the concept
of local Bayes optimality. As a side remark, we present a complete proof of a theorem by 
\citet{ICML2012Ye_175}.
\\[1ex]
\end{abstract}


\section{Introduction}
\label{se:intro}

The formal definition of quantification as a machine learning task is often credited to 
\citet{forman2008quantifying} who wrote: ``\emph{The quantification task for machine learning:} 
given a limited training set with
class labels, induce a quantifier that takes an unlabeled test set as input and returns
its best estimate of the number of cases in each class. In other words, the quantification
task is to accurately estimate the test class distribution via machine
learning, but without assuming a large training set that is sampled at random
from the test distribution. The input to a quantifier is a batch of cases, whereas a
traditional classifier takes a single case at a time and predicts its single class (or
a distribution of classes reflecting its uncertainty about that one case).''

At least since the 1960s \citep{buck1966comparison},  researchers and practitioners were aware of the need 
to track changes of the prior probabilities (or prevalences) of classes between different datasets.
In the machine learning community, the topic received renewed attention after \citet{saerens2002adjusting}
suggested a powerful alternative to the `confusion matrix method' they considered the standard
approach at the time.

\citet{fawcett2005response} marked another milestone in the discussion of how to deal with changed prevalences 
when they noticed that as a consequence of different causalities there are different dataset shift regimes 
that need to be tackled in different ways. Since then a number of papers has been published with
proposals of how to categorise different types of dataset shift \citep{storkey2009training,
MorenoTorres2012521, kullpatterns}.

There are two types of dataset shift between training and target dataset that can be easily 
characterised. Under `covariate shift', the assumption is that the posterior (i.e.\ conditional)
class probabilities are the same on the training and the target datasets. However, the
distribution of the covariates (or features) may change. This change should be taken into 
account already when a classifier is learnt  
\citep{shimodaira2000improving, sugiyama2007covariate, bickel2009discriminative}. The other
easily characterised dataset shift type is `prior probability shift' where the
feature distributions conditional on the classes remain the same when datasets are switched. Typically, in this case,
adjustments to the posterior class probabilities and decision thresholds are recommended in the literature
\citep{Elkan01,   forman2008quantifying, Xue:2009:QSC:1557019.1557117, 
hopkins2010method, bella2010quantification}. Another approach is the
direct estimation of the changed prior probabilities by minimising the distance of 
the feature distributions on the training and target datasets 
\citep{saerens2002adjusting, forman2008quantifying,  gonzalez2013class, hofer2013drift, duPlessis2014110,
kawakubo2016computationally}.

Other types of dataset shift are less easy to describe and to deal with. \citet{tasche2014exact} 
defined the `invariant density ratio' dataset shift which generalises the prior probability 
shift in such a way that only ratios of feature densities but not the densities themselves are
unchanged. A recent paper by \citet{hofer2015adapting} is outstanding by dealing with 
dataset shift under very weak assumptions on the structure of the shift.

\citet{Esuli2010} suggested that specially trained classifiers 
(called quantifiers) can be used for the so-called `Classify \& Count' quantification \citep{forman2008quantifying}
without a need of adjustments to the
estimates. Classify \& Count means that a classifier is optimised for classification on a training set and 
applied to the target dataset for the prediction of class labels. The number of
predicted positive labels (in the case of binary quantification) is then used as an estimate of the 
prevalence of the positive class in the target
dataset. The practical implementation of this proposal for binary classification 
and quantification has been explored in papers by
\citet{Milli2013}, \citet{barranquero2015quantification} and \citet{esuli2015optimizing}.
In all these three papers, emphasis is put on the need to have the quantifier properly calibrated on the 
training dataset, in the sense that the number of objects predicted to be positive should be
equal to the true number of positive objects.

The experiments by \citet{Milli2013} suggest that after all Classify \& Count with
adjustments works better than pure  Classify \& Count. In contrast,  \citet{barranquero2015quantification} 
and \citet{esuli2015optimizing} report Classify \& Count quantification performance of the specially trained 
quantifiers that is at least comparable to the performance of classifiers with adjustments.
\citet{esuli2015optimizing} are somewhat ambiguous with respect to clearly stating  all of their optimisation 
criteria. This is different to \citet{barranquero2015quantification} whose authors clearly say that the
calibration criterion must be supplemented by a condition enforcing good classification. 
This is an intuitive requirement because, in particular, the complementarity of the two criteria of calibration
and classification seems to guarantee uniqueness of the optimal quantifier.

In this paper, we discuss the theoretical foundations and limitations of quantification
without adjustments and illustrate the insights with the classical example of the binormal model with 
equal variances \citep{vanTrees}. We focus on analysis of the `Q-measure'
approach by \citet{barranquero2015quantification} 
because it has been well and intuitively documented.  Our first finding is that 
quantification without adjustments in principle may work if proper calibration on 
the training dataset is ensured and the positive class prevalences 
on the training and target datasets are the same or nearly the same. However, 
a second finding is that the Q-measure approach has to be deployed with care 
because the quantifiers resulting from it may be miscalibrated. All in all, 
these findings suggest that the potential applications of quantification without 
adjustments are rather limited.

This paper is organised as follows: In Section~\ref{se:main} and its subsections, the
theory of binary classification is revisited such that the theoretically best quantifiers can
be identified for some specific optimisation criteria. 
In particular, we introduce the concept of `local Bayes optimality' and discuss 
its applications to minimax tests of two simple hypotheses and the
optimisation of F-measure in the context of binary classification. As another application of the results 
from Section~\ref{se:main}, 
in Section~\ref{se:binormal} the
experiment of \citet{barranquero2015quantification} is revisited and reviewed in the fully controlled
setting of the binormal model with equal variances. Section~\ref{se:concl} concludes the paper.
As a side remark, we present in Appendix~\ref{se:OptClass} a complete proof of Theorem~4 of 
\citet{ICML2012Ye_175} who published only the key argument of the proof but did not mention other
important steps.


\section{Classifying for quantification}
\label{se:main}

In order to be able to appropriately assess the merits and limitations of the proposal by 
\citet{barranquero2015quantification} we adopt a precise mathematical formalism. Based on 
this formalism, we can characterise `locally optimal binary classifiers' which are
closely related to both optimal Bayes classifiers and optimal Neyman-Pearson classifiers 
\citep[][Section~2.2]{vanTrees}. The concept of local optimality allows us to 
characterise minimax tests of two simple hypotheses in a way alternative to that
by \citet[][Chapter~5]{scharf1991statistical} and to provide an alternative proof
of Theorem~4 of \citet{ICML2012Ye_175} on F-measure-optimal classifiers. In Section~\ref{se:binormal},
we use results of this section to explore the \citeauthor{barranquero2015quantification} proposal in detail
by inspecting its implementation for the binormal model with equal variances. 


\subsection{Locally optimal binary classifiers}

We discuss binary classification and the properties of classifiers in a probabilistic setting specified 
by a probability space as it was done by many authors before \citep[see, e.g.][]{vanTrees}. 
The probability space $(\Omega, \mathcal{A}, \mathrm{P})$ describes the experiment of choosing
an object at random\footnote{%
The case of $\Omega$ being finite and $\mathrm{P}$ being the uniform distribution on $\Omega$ provides the
sample-based setting which often is assumed for machine learning papers.}. 
The object has a class label and features. The features can be observed immediately while,
depending on whether the probability space is interpreted as a training sample or target sample (sometimes
also called test sample), 
the label is also observable at once or can be observed only with some delay.
We interpret $\mathcal{A}$ as the $\sigma$-field \citep[see, e.g.][Section~2]{billingsley3rd}
of all admissible events,
including events that cannot yet be observed. In addition, we have a $\sigma$-field $\mathcal{H}$
which is the family of the events that can be observed now.
The event $A$ with $A\in\mathcal{A}$ but $A\notin\mathcal{H}$ reveals the object's class label. 
If $A$ occurs the object has got class label $1$ (positive). 
If $A^c = \Omega\backslash{}A$ occurs the object's label is $-1$ (negative). 

\begin{assumption}\label{as:0}\ 
\begin{itemize}
\item $(\Omega, \mathcal{A}, \mathrm{P})$ is a probability space. This space describes the experiment
	of selecting an object from a population at random and observing its features and (typically with some
	delay) class label.
\item $A \in \mathcal{A}$ is a fixed event with $0<\mathrm{P}[A]<1$. If $A$ is observed, the object's 
	class label is 1, otherwise if $A^c = \Omega\backslash{}A$ is observed, the object's class label is -1. 
\item $\mathcal{H} \subset \mathcal{A}$ is a sub-$\sigma$-field of
$\mathcal{A}$ such that $A \notin \mathcal{H}$. $\mathcal{H}$ is the $\sigma$-field of immediately observable
events and, in particular, features.
\end{itemize}
\end{assumption}

In a binary classification problem setting, typically there are random variables $X: \Omega \to \mathbb{R}^d$
for some $d \in \mathbb{N}$ (vector of explanatory variables or scores) 
and $Y:\Omega \to \{-1, 1\}$ (dependent or class variable) such that $\mathcal{H} = \sigma(X)$ and
$Y^{-1}(\{1\}) = A$.

In the setting of Assumption~\ref{as:0}, typically one wants to predict an object's
class label (i.e.\ predict whether or not event $A$ has occurred) based on the observable information
captured in the events $H\in\mathcal{H}$. Each of these events $H$ defines a binary `classifier'
in the following sense:
\begin{itemize}
\item If $H$ occurs the object's label is predicted as $1$.
\item If $H^c = \Omega\backslash H$ occurs the object's label is predicted as -$1$.
\end{itemize}
This way binary classifiers are identified with elements of $\mathcal{H}$. We therefore do not introduce
extra notation for classifiers. Note that the object's class $A$ does not define a classifier because
by assumption we have $A\notin\mathcal{H}$.

Define the expected `misclassification cost' $L_{a,b}(H)$ for $H \in \mathcal{H}$ and fixed $a, b \ge 0$ with $a+b>0$ by
\begin{equation}\label{eq:miscl}
	L_{a,b}(H) \ = \ a\,\mathrm{P}[H^c\cap A] + b\,\mathrm{P}[H\cap A^c].
\end{equation}
According to \eqref{eq:miscl}, there is no cost for misclassification if the label is correctly predicted 
(i.e.\ if the event $(A\cap H) \cup (A^c \cap H^c)$ occurs).
If $1$ is predicted for true label $-1$ (event $H\cap A^c$, false positive), the cost is $b$. 
If $-1$ is predicted for true label $1$ (event $H^c\cap A$, false negative), the cost is $a$.
The misclassification cost $L_{a,b}(H)$ is the expected cost for the classifier represented by 
event $H$.

From Section~2.2 of \citet{vanTrees} or Section~1.3 of \citet{Elkan01}, 
we know the optimal choice $H^\ast$ (`Bayes classifier') of $H$ for minimising $L_{a,b}(H)$:
\begin{equation}\label{eq:GlobalMin}
H^\ast \ \stackrel{\mathrm{def}}{=} \ \bigl\{\mathrm{P}[A\,|\,\mathcal{H}] > \tfrac b{a+b}\bigr\} \ = \
\bigl\{\omega \in \Omega: \mathrm{P}[A\,|\,\mathcal{H}](\omega) > \tfrac b{a+b}\bigr\} \ = \
\arg \underset{H\in\mathcal{H}}{\min} L_{a,b}(H),
\end{equation}
where $\mathrm{P}[A\,|\,\mathcal{H}]$ denotes the conditional probability (or posterior probability)
 of $A$ given $\mathcal{H}$ as 
defined in standard text books on probability theory \citep[e.g.][Section~33]{billingsley3rd}. The
following proposition shows that in some sense all classifiers of the shape 
\begin{equation}\label{eq:shape}
H_q \ \stackrel{\mathrm{def}}{=}\ \{\mathrm{P}[A\,|\,\mathcal{H}] > q\}, \quad 0 < q < 1,
\end{equation}
are local minimisers of $L_{a,b}(H)$.

\begin{proposition}\label{pr:local} Under Assumption~\ref{as:0}, 
let $a, b \ge 0$ with $a+b>0$ be fixed. Define $L_{a,b}(H)$ and $H_q$ by 
\eqref{eq:miscl} and \eqref{eq:shape} respectively. Let $p_q = \mathrm{P}[H_q]$. Then the following
two statements hold:
\begin{itemize}
\item[(i)] $q < \frac b{a+b}$ \quad$\Rightarrow$ \quad
$H_q = \arg \underset{H\in\mathcal{H},\,\mathrm{P}[H] \ge p_q}{\min} L_{a,b}(H)$.
\item[(ii)] $q > \frac b{a+b}$ \quad$\Rightarrow$ \quad
$H_q = \arg \underset{H\in\mathcal{H},\,\mathrm{P}[H] \le p_q}{\min} L_{a,b}(H)$.
\end{itemize}
\end{proposition}
\textbf{Proof.} Let $H \in \mathcal{H}$ be given. With some algebra\footnote{%
$\mathbf{1}_S$ denotes the indicator function of the set $S$, i.e.\ $\mathbf{1}_S(s)=1$ for $s\in S$ and
$\mathbf{1}_S(s)=0$ for $s\notin S$.}, it can be shown that 
\begin{align}
L_{a,b}(H) & =  a\,\mathrm{P}[A] + \bigl(b-(a+b)\,q\bigr)\,\mathrm{P}[H] + 
	(a+b)\,\mathrm{E}\bigl[(q-\mathrm{P}[A\,|\,\mathcal{H}])\,\mathbf{1}_{H\cap H_q}\bigr]\notag \\
	& \qquad + (a+b)\,\mathrm{E}\bigl[(q-\mathrm{P}[A\,|\,\mathcal{H}])\,
			\mathbf{1}_{H\cap H^c_q}\bigr]\notag\\
			& \ge a\,\mathrm{P}[A] + \bigl(b-(a+b)\,q\bigr)\,\mathrm{P}[H] + 
	(a+b)\,\mathrm{E}\bigl[(q-\mathrm{P}[A\,|\,\mathcal{H}])\,\mathbf{1}_{H\cap H_q}\bigr]\notag\\
& \ge a\,\mathrm{P}[A] + \bigl(b-(a+b)\,q\bigr)\,\mathrm{P}[H] + 
	(a+b)\,\mathrm{E}\bigl[(q-\mathrm{P}[A\,|\,\mathcal{H}])\,\mathbf{1}_{H_q}\bigr].\label{eq:algebra}
\end{align}
In case $q < \frac b{a+b}$ we have $b-(a+b)\,q > 0$. Then  it holds that
$\bigl(b-(a+b)\,q\bigr)\,\mathrm{P}[H] \ge \bigl(b-(a+b)\,q\bigr)\,p_q$ for $\mathrm{P}[H] \ge p_q$.
By \eqref{eq:algebra}, this implies (i).

In case $q > \frac b{a+b}$ we have $b-(a+b)\,q < 0$. Then  it holds that
$\bigl(b-(a+b)\,q\bigr)\,\mathrm{P}[H] \ge \bigl(b-(a+b)\,q\bigr)\,p_q$ for $\mathrm{P}[H] \le p_q$.
From this observation and \eqref{eq:algebra}, statement (ii) follows. \hfill \qed

Proposition~\ref{pr:local} is about 'locally' optimal classifiers in the sense that only classifiers
with identical probability of predicting $1$ are compared. We state this observation more
precisely in item (i) of the following remark:

\begin{remark}\label{rm:comments}\ 
\begin{itemize}
\item[(i)] We have $H^\ast = H_{\frac b{a+b}}$ for $H^\ast$ as defined in \eqref{eq:GlobalMin}. 
Hence the case $q = \frac b{a+b}$ is not treated in Proposition~\ref{pr:local} because it is covered by
\eqref{eq:GlobalMin}. Nonetheless, it is worth noting that Proposition~\ref{pr:local} and 
\eqref{eq:GlobalMin} together imply that for all\/
$a, b \ge 0$ with $a+b > 0$ and $0 < q < 1$ it holds that
\begin{equation*}
H_q \ =\ \arg \underset{H\in\mathcal{H},\,\mathrm{P}[H] = p_q}{\min} L_{a,b}(H).
\end{equation*}
\item[(ii)] In the case $a = 0$, $b = (1-\mathrm{P}[A])^{-1}$, Proposition~\ref{pr:local}~(i) implies for 
all $0 < q < 1$ that
\begin{equation*}\label{eq:fpr}
	H_q \ =\ \arg \underset{H\in\mathcal{H},\,\mathrm{P}[H] \ge p_q}{\min} 
		\frac{\mathrm{P}[H \cap A^c]}{\mathrm{P}[A^c]} \ = \ 
		\arg \underset{H\in\mathcal{H},\,\mathrm{P}[H] \ge p_q}{\min} \mathrm{P}[H \,|\, A^c].
\end{equation*}
$\mathrm{P}[H \,|\, A^c]$ is called `false positive rate' (FPR). 
\item[(iii)] In the case $a = \mathrm{P}[A]^{-1}$ and $b = 0$, Proposition~\ref{pr:local}~(ii) implies for 
all\/ $0 < q < 1$ that
\begin{equation*}\label{eq:tpr}
	H_q \ =\ \arg \underset{H\in\mathcal{H},\,\mathrm{P}[H] \le p_q}{\max} 
		\frac{\mathrm{P}[H \cap A]}{\mathrm{P}[A]} \ = \ 
		\arg \underset{H\in\mathcal{H},\,\mathrm{P}[H] \le p_q}{\max} \mathrm{P}[H \,|\, A].
\end{equation*}
$\mathrm{P}[H \,|\, A]$ is called `true positive rate' (TPR).
\end{itemize}
\end{remark}

As mentioned above, Proposition~\ref{pr:local} may be interpreted as a result in between the 
characterisation of optimal Bayes classifiers and the optimal classifier (test) from the Neyman-Pearson 
lemma. The following theorem gives a precise statement of this observation.

\begin{theorem}\label{th:max}
Let $(\Omega_0, \mathcal{M}, \mu)$ be a measure space. 
Assume that $\mathrm{Q}^-$ and $\mathrm{Q}^+$ are probability measures on 
$(\Omega_0, \mathcal{M})$ 
which are absolutely continuous with respect to $\mu$. Assume furthermore that the 
densities $f^-$ and $f^+$ of\/
$\mathrm{Q}^-$ and\/ $\mathrm{Q}^+$ respectively are positive. Define the likelihood ratio $\lambda$ by $\lambda = \frac{f^+}{f^-} >0$.
If the distribution of $\lambda$ is continuous under $\mathrm{Q}^-$ and $\mathrm{Q}^+$, i.e.\ $\mathrm{Q}^-[\lambda=\ell]=0=\mathrm{Q}^+[\lambda=\ell]$
for all\/ $\ell > 0$, then there is a number $\ell^\ast >0$ with $\mathrm{Q}^-[\lambda > \ell^\ast] = \mathrm{Q}^+[\lambda \le \ell^\ast]$
such that 
\begin{equation}\label{eq:crit}
    \underset{M\in\mathcal{M}}{\min} \max\bigl(\mathrm{Q}^-[M], \mathrm{Q}^+[M^c]\bigr)\ = \
    \mathrm{Q}^-[\lambda > \ell^\ast].
\end{equation}
\end{theorem}
\textbf{Proof.} Define the probability space $(\Omega, \mathcal{A}, \mathrm{P})$ by 
\begin{itemize}
\item $\Omega = \Omega_0 \times \{-1, 1\}$  with
projections $X(\omega, c) = \omega$ and $Y(\omega, c) = c$ for $(\omega, c) \in \Omega$, 
\item $\mathcal{A} = \mathcal{M} \otimes \mathcal{P}(\{-1, 1\}) = \sigma(X, Y)$, 
\item $\mathrm{P}[Y = -1] = 1/2 = \mathrm{P}[Y = 1]$ as well as  $\mathrm{P}[X \in M\,|\,Y=-1] = \mathrm{Q}^-[M]$ and
    $\mathrm{P}[X \in M\,|\,Y=1] = \mathrm{Q}^+[M]$ for $M\in\mathcal{M}$.
\end{itemize}
Then Assumption~\ref{as:0} is satisfied if $A$ is chosen as $A = \{Y = 1\}$ and $\mathcal{H}$ is chosen as
$\mathcal{H} = \sigma(X) = \mathcal{M} \times \bigl\{\emptyset, \{-1, 1\}\bigr\}$. By construction of $\mathrm{P}$,
it follows that the probability of $A=\{Y=1\}$ conditional on $\mathcal{H}$ is given by
\begin{equation*}
    \mathrm{P}\bigl[Y=1\,|\,\mathcal{H}\bigr] \ = \ \frac{f^+\circ X}{f^+\circ X + f^-\circ X}.
\end{equation*}
This implies for any $0 < q < 1$ that
\begin{equation}\label{eq:rep}
    \mathrm{P}\bigl[\mathrm{P}[Y=1\,|\,\mathcal{H}] > q\bigr] \ = \
    \frac{\mathrm{Q}^-[\lambda > \frac{q}{1-q}]+\mathrm{Q}^+[\lambda > \frac{q}{1-q}]}{2}.
\end{equation}
Fix $M\in \mathcal{M}$ and let $H = \{X\in M\}$. The assumption on the continuity of the distribution of $\lambda$ under
$\mathrm{Q}^-$ and $\mathrm{Q}^+$ implies that 
$$
\underset{M\in\mathcal{M}}{\min} \max\bigl(\mathrm{Q}^-[M], \mathrm{Q}^+[M^c]\bigr) \ < \ 1.
$$  
For this proof, we therefore may assume without loss of generality 
that $\max\bigl(\mathrm{Q}^-[M], \mathrm{Q}^+[M^c]\bigr) < 1$ and hence
$0<\mathrm{P}[H]<1$. Again by the assumption on the continuity of the distribution of $\lambda$
under $\mathrm{Q}^-$ and $\mathrm{Q}^+$, then \eqref{eq:rep} implies that there is $q = q(H)$ such that
$\mathrm{P}[H] = \mathrm{P}\bigl[\mathrm{P}[Y=1\,|\,\mathcal{H}] > q\bigr]$. From Remark~\ref{rm:comments} (ii) and (iii)
now it follows that 
\begin{align*}
    \max\bigl(\mathrm{Q}^-[M], \mathrm{Q}^+[M^c]\bigr) &= \max\bigl(\mathrm{P}[H\,|\,Y=-1],\, \mathrm{P}[H^c\,|\,Y=1]\bigr)\\
    & \ge \max\bigl(\mathrm{P}\bigl[\mathrm{P}[Y=1\,|\,\mathcal{H}] > q\,|\,Y=-1\bigr],\, 
    \mathrm{P}\bigl[\{\mathrm{P}[Y=1\,|\,\mathcal{H}] > q\}^c\,|\,Y=1\bigr]\bigr)\\ 
    & = \max\bigl(\mathrm{Q}^-\bigl[\lambda > \tfrac{q}{1-q}\bigr],\, \mathrm{Q}^+\bigl[\lambda \le \tfrac{q}{1-q}\bigr]\bigr).
\end{align*}
Since this holds for all\/ $M\in \mathcal{M}$ with $\max\bigl(\mathrm{Q}^-[M], \mathrm{Q}^+[M^c]\bigr) < 1$, we can conclude that
$$
\underset{M\in\mathcal{M}}{\min} \max\bigl(\mathrm{Q}^-[M], \mathrm{Q}^+[M^c]\bigr) \ \ge \
\underset{\ell > 0}{\min} \max\bigl(\mathrm{Q}^-\bigl[\lambda > \ell\bigr],\, \mathrm{Q}^+\bigl[\lambda \le \ell\bigr]\bigr).
$$ 
The intermediate value theorem implies that there is an $\ell^\ast$ such that 
$\mathrm{Q}^-[\lambda > \ell^\ast] = \mathrm{Q}^+[\lambda \le \ell^\ast]$. Since 
$\ell \mapsto \mathrm{Q}^-\bigl[\lambda > \ell\bigr]$ is non-increasing and  
$\ell \mapsto \mathrm{Q}^+\bigl[\lambda \le \ell\bigr]$ is non-decreasing, it follows that\\[1ex]
\hspace*{3cm}$\displaystyle{}\underset{\ell > 0}{\min} \max\bigl(\mathrm{Q}^-\bigl[\lambda > 
\ell\bigr],\, \mathrm{Q}^+\bigl[\lambda \le \ell\bigr]\bigr) 
\ = \ \mathrm{Q}^-[\lambda > \ell^\ast]$. \hfill \qed

\begin{remark}\label{rm:thcomments} One interpretation of Theorem~\ref{th:max} is as providing a 
minimax test for the decision between two simple hypotheses  
\citep[see Chapter~5 of][]{scharf1991statistical}. 
The test problem is to distinguish $\mathrm{Q}^-$ and
$\mathrm{Q}^+$. Tests are characterised by observable sets $M\in\mathcal{M}$ where $\omega \in M$ means
'accept $\mathrm{Q}^+$' and $\omega \notin M$ means 'reject  $\mathrm{Q}^+$ in favour of  $\mathrm{Q}^-$'.
However, in contrast to the setting of the Neyman-Pearson lemma, 
none of the two hypotheses is considered more important than the other. Therefore, as expressed on 
the left-hand side of \eqref{eq:crit}, an optimal test is meant to minimise the probabilities of the
type I and II errors at the same time.

\eqref{eq:crit} shows that in a 'continuous' setting there is an optimal test under the criterion 
given by the left-hand side of \eqref{eq:crit}
which is based on the likelihood ratio $\lambda$, i.e.\ the ratio of the densities of the tested probability
measures. Hence the structure of the optimal test is the same as for the optimal Neyman-Pearson test, the 
cost-optimal Bayes test and the minimax optimal Bayes test (see Section~2.2 of \citealp{vanTrees}, or
Chapter~5 of \citealp{scharf1991statistical}).
\end{remark}

The concept of local Bayes optimality (i.e.~Proposition~\ref{pr:local}) can also be applied to the 
question of how to determine binary classifiers that are optimal with respect to the `F-measure' criterion.
$F_\beta$-measure for $\beta > 0$ was introducted by \citet{van1974foundation} in order to avoid neglecting 
the minority class when learning binary classifiers. For a given classifier, its $F_\beta$-measure is defined as
$$F_\beta\ =\ \frac{1+\beta^2}{\frac{\beta^2}{\mathrm{Recall}}+\frac 1{\mathrm{Precision}}},$$
where the classifier's `precision' is the ratio of the number of positively predicted true positive objects and
the number of positively predicted objects while its `recall' is the ratio of the number of positively predicted 
true positive objects and the number of true positive objects.

In the setting of Assumption~\ref{as:0}, we have
$$\mathrm{Precision} \ = \ \mathrm{P}[A\,|\,H] \quad \text{and}\quad \mathrm{Recall} \ = \ \mathrm{P}[H\,|\,A],$$
where $A$ denotes `positive class' and $H$ stands for `classifier predicts positive'. Note that recall is 
identical with `true positive rate' as defined in Remark~\ref{rm:comments}.
We can rewrite the definition of $F_\beta$ as 
\begin{equation}\label{eq:Fbeta}
F_\beta(H)\ =\ \frac{\beta^2+1}{\frac{\beta^2}{\mathrm{P}[H\,|\,A]} + 
    \frac{1}{\mathrm{P}[A\,|\,H]}}
= \frac{(1+\beta^2)\,\mathrm{P}[H \,|\, A]}{\beta^2 + 
    \mathrm{P}[H] / \mathrm{P}[A]},
\end{equation}
for $H \in \mathcal{H}$ and fixed $\beta > 0$. \citet[][Theorem~4]{ICML2012Ye_175} observed that
classifiers that are optimal in the sense of maximising $F_\beta$ can be constructed by `thresholding' 
the conditional class probability $\mathrm{P}[A\,|\,\mathcal{H}]$. In the notation of this paper,
their observation can be precisely stated  as follows:
\begin{equation}\label{eq:Fgen}
\underset{H\in\mathcal{H}}{\sup} F_\beta(H)\  = \ 
	\underset{0 \le q \le 1}{\sup} \max\bigl(F_\beta(\{\mathrm{P}[A\,|\,\mathcal{H}] > q\}), 
	F_\beta(\{\mathrm{P}[A\,|\,\mathcal{H}] \ge q\})\bigr).
\end{equation}
\citet{ICML2012Ye_175} published only the most important part of the proof\footnote{%
The proof is available in an appendix to the paper of 
\citeauthor{ICML2012Ye_175} which can be downloaded at\\
\href{https://www.comp.nus.edu.sg/~leews/publications/fscore-appendix.pdf}
{\texttt{https://www.comp.nus.edu.sg/$\sim$leews/publications/fscore-appendix.pdf}}} of \eqref{eq:Fgen}. In
Appendix~\ref{se:OptClass} of this paper, we provide a complete  proof of \eqref{eq:Fgen}. 

However, 
in the case where the conditional class probability $\mathrm{P}[A\,|\,\mathcal{H}]$ has a continuous 
distribution, \eqref{eq:Fgen} is an immediate consequence of Remark~\ref{rm:comments}~(iii): 
\begin{itemize}
\item By continuity of the distribution of $\mathrm{P}[A\,|\,\mathcal{H}]$,
for any classifier $H\in\mathcal{H}$ with $0 < \mathrm{P}[H] < 1$ there is a number $0 < q < 1$ such
that $\mathrm{P}[H] = \mathrm{P}[H_q]$ with $H_q$ defined as in \eqref{eq:shape}.
\item Therefore, Remark~\ref{rm:comments}~(iii) implies that
$$F_\beta(H)\ 
     \le\ \frac{(1+\beta^2)\,\mathrm{P}[H_q \,|\, A]}{\beta^2 + \mathrm{P}[H] / \mathrm{P}[A]}\ =\ 
     \frac{(1+\beta^2)\,\mathrm{P}[H_q \,|\, A]}{\beta^2 + \mathrm{P}[H_q] / \mathrm{P}[A]}\ =\ F_\beta(H_{q}).$$
\end{itemize}


\subsection{Application to quantification under prior probability shift}
\label{se:PriorShift}

In contrast to  \citet{esuli2015optimizing} and \citet{Milli2013}, \citet{barranquero2015quantification}
specify the dataset shift problem they are going to tackle with their proposal: It is prior probability 
shift. We modify Assumption~\ref{as:0} accordingly.

\begin{assumption}\label{as:extend}
We extend the setting of Assumption~\ref{as:0} by assuming that there is a second probability measure
$\mathrm{P}_1$ on $(\Omega, \mathcal{A})$.
$\mathrm{P}_1$ evolves from $\mathrm{P}$ by `prior probability shift', i.e.\ the
probabilities of sets $H\in\mathcal{H}$ conditional on $A$ and the
probabilities of sets $H\in\mathcal{H}$ conditional on
$A^c$ are the same under 
$\mathrm{P}$ and $\mathrm{P_1}$\/:
$$
\mathrm{P}[H\,|\,A] = \mathrm{P}_1[H\,|\,A]\quad
\text{and}\quad \mathrm{P}[H\,|\,A^c] = \mathrm{P}_1[H\,|\,A^c], \quad \text{for all}\ H\in\mathcal{H}.
$$
\end{assumption}

Under Assumption~\ref{as:extend}, we can describe for any  classifier $H\in \mathcal{H}$ the
probability $\mathrm{P}_1[H]$ as an affine function of $w = \mathrm{P}_1[A]$:
\begin{equation}\label{eq:affine}
    \mathrm{P}_1[H] \ = \ w\,\bigl(\mathrm{P}[H\,|\,A] - \mathrm{P}[H\,|\,A^c]\bigr) + \mathrm{P}[H\,|\,A^c].
\end{equation} 
In practice, the true positive rate (TPR) $\mathrm{P}[H\,|\,A]$ and the false positive rate (FPR) $\mathrm{P}[H\,|\,A^c]$
can be estimated (possibly with large potential bias) from the training set (in our setting:
$(\Omega, \mathcal{A}, \mathrm{P})$) and the probability 
$\mathrm{P}_1[H]$ of an object to be classified as positive (after prior probability shift) can be 
estimated from the target set (in our setting:
$(\Omega, \mathcal{A}, \mathrm{P}_1)$).
 Then \eqref{eq:affine} can be solved for $w= \mathrm{P}_1[A]$ to obtain an estimate of
the new prior probability of the positive class:
\begin{equation}\label{eq:confusion}
\mathrm{P}_1[A] \ = \ \frac{\mathrm{P}_1[H]-\mathrm{P}[H\,|\,A^c]}
    {\mathrm{P}[H\,|\,A]-\mathrm{P}[H\,|\,A^c]}.
\end{equation}
This approach is called `confusion matrix method' \citep{saerens2002adjusting}.
It has also been described as 
`Adjusted Count' approach \citep{forman2008quantifying} and has been deployed
by practitioners at least since the 1960s \citep{buck1966comparison}.

In theory, for any classifier $H$, \eqref{eq:confusion} provides the adjustment needed to obtain an accurate estimate of
the probability of the positive class from a potentially quite inaccurate estimate by $\mathrm{P}_1[H]$. 
Experiments by some research teams, however, have cast doubt on the appropriateness of this approach.
Both good \citep{Xue:2009:QSC:1557019.1557117, hopkins2010method} and unsatisfactory performance 
\citep{saerens2002adjusting, forman2008quantifying} of the confusion matrix method have been reported. Other
papers report mixed findings  
\citep{bella2010quantification, gonzalez2013class, hofer2013drift, duPlessis2014110}.

As \eqref{eq:confusion} is valid only under prior probability shift (Assumption~\ref{as:extend}), performance
issues with the confusion matrix method should not be a surprise in circumstances when there is little 
evidence of prior probability shift against other types of dataset shift \citep[see][for a 
taxonomy of dataset shift types]{MorenoTorres2012521}. But most if not all of the above-mentioned reports on
the performance of the confusion matrix method refer to controlled environments with prior probability shift.
A number of reasons have been identified to potentially negatively impact the confusion matrix method performance.
Among them are class imbalance in the training set \citep{forman2008quantifying} and issues with the
accurate estimation of TPR and FPR on the training set \citep{esuli2015optimizing}.

Although many other approaches to prior probability estimation have been proposed 
\citep{hofer2013drift, duPlessis2014110, hofer2015adapting, kawakubo2016computationally}, 
no gold standard has yet
emerged because all approaches appear to suffer from numerical problems to some extent.

This observation has led some authors to suggest that so-called `quantifiers' (classifiers, 
specifically developed for quantification) might be a viable
solution \citep{Esuli2010, Milli2013, barranquero2015quantification, esuli2015optimizing}.
In the notation of this paper, both classifiers and quantifiers are characterised by observable events $H\in \mathcal{H}$
which are interpreted as 'predict positive'. The difference between the concepts of 
'classifier' and 'quantifier' is the intended use, as explained in the quotation from
\citet{forman2008quantifying} in Section~\ref{se:intro}:
\begin{itemize}
    \item Classifiers are deployed for predicting the class labels of single objects. Therefore, development
    of a classifier typically involves minimising the expected loss of decisions about single objects (see,
    for instance, the right-hand-side of \eqref{eq:GlobalMin}).
    \item Quantifiers are deployed for estimating the prevalence of a class in a sample or population. 
    \citet{barranquero2015quantification} have argued that this 
    different purpose
    should be reflected in a different objective function for the development of a quantifier. They 
    suggest that with 
    an appropriate objective function, no adjustment like \eqref{eq:confusion} would be needed.
    \item In their paper, \citet{barranquero2015quantification} suggest maximising a Q-measure 
    criterion (see \eqref{eq:Qbeta} below).   
    Their experiments were  conducted in a prior probability shift setting (see Assumption~\ref{as:extend}).
    Similarly, \citet{Milli2013} report experimental results from a prior probability shift setting.
    \item As a different approach, \citet{Esuli2010} 
    suggest minimising the Kullback-Leibler distance between the
    observed class distribution and the predicted class distribution on the training set. Implicitly,
    as they work with Support Vector Machines \citep{esuli2015optimizing}, they also apply some 
    classification optimisation criterion.
    \citeauthor{esuli2015optimizing} only use 'natural' datasets such that their datashift environment cannot
    be characterised as prior probability shift.    
\end{itemize}
In the following, we focus on the analysis of the approach proposed by \citet{barranquero2015quantification} 
to deal with prior probability shift. Analysis of the approach followed by \citet{esuli2015optimizing}
is harder because \citeauthor{esuli2015optimizing} do not specify their dataset shift assumption and
because the performance of their approach seems to depend on their choice of the classifier development
methodology (as support vector machines). Analysis and potential criticism of \citet{esuli2015optimizing},
therefore, is not undertaken in this paper. The results of \citet{Milli2013} are less controversial
than those of \citet{barranquero2015quantification} and \citet{esuli2015optimizing} because
\citeauthor{Milli2013} report superior quantification performance for adjusted classifiers.

\begin{figure}[t!p]
\caption{Illustration of prediction error function~\eqref{eq:error}.}
\label{fig:PredictionError}
\begin{center}
\ifpdf
	\resizebox{\height}{12cm}{\includegraphics[width=14cm]{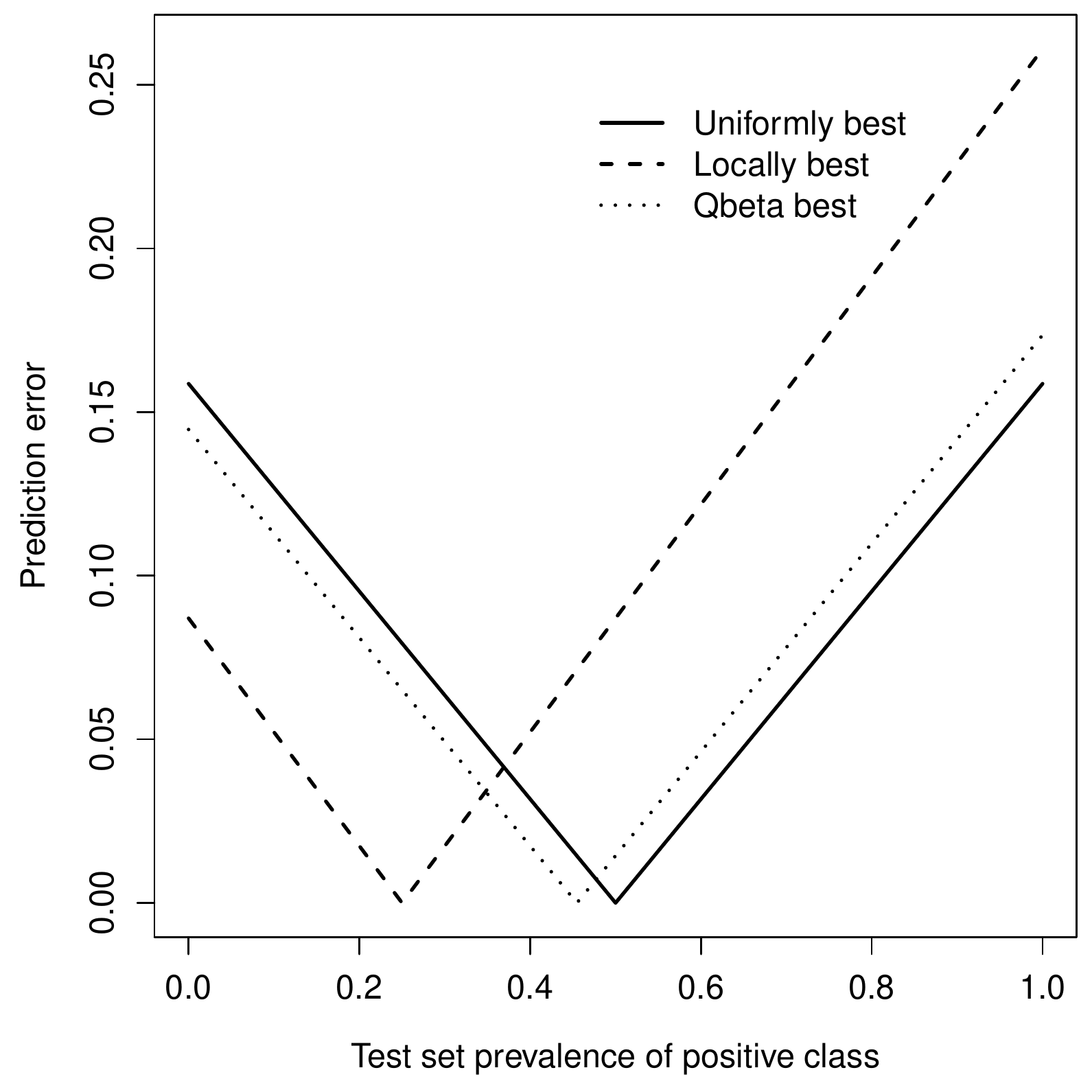}}
\fi
\end{center}
\end{figure}

For a fixed, unadjusted classifier specified by a set $H\in\mathcal{H}$, the absolute prediction error
$|\mathrm{P}_1[A]-\mathrm{P}_1[H]|$ of the prevalence of the positive class in the target dataset is
the combination of a decreasing and an increasing straight line if it is represented as a function
of the true positive class prior probability $w = \mathrm{P}_1[A]$:
\begin{equation}\label{eq:error}
	\mathcal{E}_H(w) \ = \ 
	\begin{cases}
	w\,\bigl(\mathrm{P}[H\,|\,A] - \mathrm{P}[H\,|\,A^c]-1\bigr) + \mathrm{P}[H\,|\,A^c], & 
	\text{for}\ w \le \frac{\mathrm{P}[H\,|\,A^c]}{\mathrm{P}[H\,|\,A^c] + 1-\mathrm{P}[H\,|\,A]},\\
	w\,\bigl(1-\mathrm{P}[H\,|\,A] +\mathrm{P}[H\,|\,A^c]\bigr) - \mathrm{P}[H\,|\,A^c], 
	& \text{otherwise}.
	\end{cases}
\end{equation}
See Figure~\ref{fig:PredictionError} for an illustration of the absolute prediction error concept. The
rationale for the three different curves is explained in Section~\ref{se:binormal} below. For the moment,
if we ignore the question of how to minimise the absolute prediction error, the figure tells us that
every classifier is a perfect predictor of one positive class prevalence in the target dataset.
Unfortunately, it is not very helpful to know this because it is perfection in the way a broken clock
is perfectly right once a day. Hence it is worthwhile to try and find out more about 
minimising the error as we do in the following.

\eqref{eq:error} immediately implies the following result on error bounds for the prediction of the
positive class prevalence by a classifier:
\begin{proposition}\label{pr:error}
Under Assumption~\ref{as:extend}, the following inequality holds for any $H\in \mathcal{H}$:
\begin{equation}
\bigl|\mathrm{P}_1[A] - \mathrm{P}_1[H]\bigr| \ \le \ \max\bigl(\mathrm{P}[H\,|\,A^c], \,
	\mathrm{P}[H^c\,|\,A]\bigr).
\end{equation}
\end{proposition}
Proposition~\ref{pr:error} shows that a classifier's prediction error with regard to the positive class 
prevalence is controlled by the classifier's false positive rate (FPR) $\mathrm{P}[H\,|\,A^c]$ and
its false negative rate (FNR) $\mathrm{P}[H^c\,|\,A]$. Theorem~\ref{th:max} (or Remark~\ref{rm:comments}
(ii) and (iii)) makes it possible to 
identify the optimal (in this case the minimax) classifier with regard to the prediction of
the positive class prevalence:

\begin{corollary}\label{co:best}
Under Assumption~\ref{as:extend}, define the classifiers $H_q$, $0 < q < 1$ by \eqref{eq:shape}.
Assume in addition, that the distribution of\/ $\mathrm{P}[A\,|\,\mathcal{H}]$ is continuous under
both $\mathrm{P}[\cdot\,|\,A]$ and $\mathrm{P}[\cdot\,|\,A^c]$. Then there is a number $0<q^\ast<1$ with
$\mathrm{P}[H_{q^\ast}\,|\,A^c] = 1- \mathrm{P}[H_{q^\ast}\,|\,A]$ such that
$$
	\underset{H\in\mathcal{H}}{\min}\max\bigl(\mathrm{P}[H\,|\,A^c], \,
	\mathrm{P}[H^c\,|\,A]\bigr)\ = \ \mathrm{P}[H_{q^\ast}\,|\,A^c].
$$
\end{corollary}
Corollary~\ref{co:best} is nice in telling us which classifier 
minimises at the same time the probabilities of `false negative' and `false positive' predictions  
on the target  dataset for whatever value of the positive class prevalence. 
Note the similarity between this classifier and the classifier serving as the basis for the 'method X' 
of \citet{forman2008quantifying}. It is also interesting to see that  
'Method Max' of \citet{forman2008quantifying} (maximise TPR - FPR) is a special case
of \eqref{eq:GlobalMin} with $a = 1/\mathrm{P}[A]$, $b=1/(1-\mathrm{P}[A])$.

But from \eqref{eq:error} it follows that the prediction error $\mathcal{E}_{H_{q^\ast}}(w)$ of $H_{q^\ast}$ 
is zero if and only if $w = 1/2$, i.e.\ if the prior positive class probability on the target dataset
is 50\%. This may seem unsatisfactory in particular if the positive class prevalence $\mathrm{P}[A]$ on the
training set is very different from 1/2. It might be more appropriate to have prediction error zero 
for $w = \mathrm{P}[A]$, i.e.\ if the positive class prevalence on the target dataset is the same as on the
training set. For this case the following result applies:
\begin{corollary}\label{co:localbest}
Under Assumption~\ref{as:extend}, define the classifiers $H_q$, $0 < q < 1$ by \eqref{eq:shape}.
Assume in addition, that there is a number $0<r<1$ such that $\mathrm{P}[H_r] = \mathrm{P}[A]$.
Then it holds that
$$
	\underset{H\in\mathcal{H},\,\mathrm{P}[H]=\mathrm{P}[A]}{\min}\max\bigl(\mathrm{P}[H\,|\,A^c], \,
	\mathrm{P}[H^c\,|\,A]\bigr)\ = \ \max\bigl(\mathrm{P}[H_r\,|\,A^c], \,
	\mathrm{P}[H_r^c\,|\,A]\bigr).
$$
\end{corollary}
It can be easily checked that for $H_r$ from Corollary~\ref{co:localbest} we have 
$\mathcal{E}_{H_r}(\mathrm{P}[A]) = 0$, with $\mathcal{E}$ defined as in \eqref{eq:error}.


\subsection{The Q-measure approach}


\citet{barranquero2015quantification} define the `Normalized Absolute Score' (NAS) for measuring 
how well a classifier (characterised by a set $H \in \mathcal{H}$, as explained below Assumption~\ref{as:0})
predicts the prior class probabilities $\mathrm{P}[A]$ and $1-\mathrm{P}[A]$ in the binary 
classification setting as described by Assumption~\ref{as:0}:
\begin{subequations}
\begin{equation}
\mathrm{NAS}(H) \ \stackrel{\textrm{def}}{=} \ 1 - \frac{|\mathrm{P}[H]-\mathrm{P}[A]|}{\max(\mathrm{P}[A], 1-\mathrm{P}[A])}.
\end{equation}
By definition, we have $\mathrm{NAS}(H) = 1$ if and only if $\mathrm{P}[H] = \mathrm{P}[A]$.
Otherwise, the range of $\mathrm{NAS}(H)$ depends on the value of $\mathrm{P}[A]$. If $\mathrm{P}[A] 
\le 1 - \mathrm{P}[A]$, then $\mathrm{P}[H] = 0$ implies $\mathrm{NAS}(H) = \frac{1-2\,\mathrm{P}[A]}{1-\mathrm{P}[A]}$
and $\mathrm{P}[H] = 1$ implies $\mathrm{NAS}(H) = 0$. If $\mathrm{P}[A] 
> 1 - \mathrm{P}[A]$, then $\mathrm{P}[H] = 0$ implies $\mathrm{NAS}(H) = 0$ and 
and $\mathrm{P}[H] = 1$ implies $\mathrm{NAS}(H) = \frac{2\,\mathrm{P}[A]-1}{\mathrm{P}[A]}$. 

The dependence of the range  of $\mathrm{NAS}$ on the value of $\mathrm{P}[A]$ is unsatisfactory because it makes
comparison of $\mathrm{NAS}$ values computed for different underlying values of $\mathrm{P}[A]$ incommensurable and,
potentially, could entail bias if $\mathrm{NAS}$ is used as an optimization criterion.
The following alternative definition of $\mathrm{NAS}^\ast$ avoids these issues:
\begin{equation}\label{eq:NAS.star}
\mathrm{NAS}^\ast(H) \ \stackrel{\textrm{def}}{=} \ 1 - \frac{\max(\mathrm{P}[H]-\mathrm{P}[A],0)}{1-\mathrm{P}[A]}
     - \frac{\max(\mathrm{P}[A]-\mathrm{P}[H],0)}{\mathrm{P}[A]}.
\end{equation}
\end{subequations}
By this definition, we have $\mathrm{NAS}^\ast(H) = 1$ if and only if $\mathrm{P}[H] = \mathrm{P}[A]$,
$\mathrm{NAS}^\ast(H) = 0$ if and only if $\mathrm{P}[H] \in \{0,1\}$, and 
$0 < \mathrm{NAS}^\ast(H) < 1$ otherwise. In the following, we use $\mathrm{NAS}^\ast$ instead of
$\mathrm{NAS}$ in order to make sure that the full potential of the approach by 
\citet{barranquero2015quantification} is realised.

\citet{barranquero2015quantification} suggest training `reliable' classifiers in order to
predict prior (unconditional) class probabilities on the target dataset under prior probability shift.
With reliability they mean that the classifiers in question should perform well in terms 
of being good at classification and being good at quantification at the same time.

In order to train an optimal quantifier for predicting the class probability $\mathrm{P}[A]$, 
\citet{barranquero2015quantification} suggest to maximise for some $\beta > 0$ the `$Q_\beta$-measure' over all possible 
classifiers (characterised by the sets $H \in \mathcal{H}$ whose outcomes trigger the prediction of class~$1$):
\begin{subequations}
\begin{align}\label{eq:Qbeta}
Q_\beta(H) &\ \stackrel{\textrm{def}}{=} \ \frac{(1+\beta^2)\,\mathrm{P}[H\,|\,A]\,\mathrm{NAS}(H)}
    {\beta^2\,\mathrm{P}[H\,|\,A] + \mathrm{NAS}(H)}\\
    &\ = \ \frac{1+\beta^2}
    {\frac{\beta^2}{\mathrm{NAS}(H)} + \frac1{\mathrm{P}[H\,|\,A]}}.\label{eq:Qalter}
\end{align}
\end{subequations}
By the definitions of $\mathrm{P}[H\,|\,A]$ and $\mathrm{NAS}(H)$ the denominator on the right-hand side
of \eqref{eq:Qbeta} takes the value $0$ if and only if $\mathrm{P}[H] = 0$. Representation \eqref{eq:Qalter} of
$Q_\beta(H)$ implies $\lim_{\mathrm{P}[H]\to 0} Q_\beta(H) = 0$. Therefore we can define $Q_\beta(H)$  by
\eqref{eq:Qbeta} for $H\in \mathcal{H}$ with $\mathrm{P}[H] > 0$ and $Q_\beta(H)=0$ for
$H\in \mathcal{H}$ with $\mathrm{P}[H] = 0$.

$Q_\beta(H)$ is a weighted harmonic mean of the true positive rate $\mathrm{P}[H\,|\,A]$ and the normalized
absolute score and, as such, is increasing in both $\mathrm{P}[H\,|\,A]$ and $\mathrm{NAS}(H)$.
\citet{barranquero2015quantification} suggest that by maximising $Q_\beta(H)$ over $H$ for fixed $A$ on a
training set,
the resulting classifier should be able to provide good estimates of $\mathrm{P}_1[A]$ on target datasets with
possibly different prior class distributions.

\begin{figure}[t!p]
\caption{Illustration of Proposition~\ref{pr:alter} and Corollary~\ref{co:alter}.}
\label{fig:Qbeta}
\begin{center}
\ifpdf
	\resizebox{\height}{12.0cm}{\includegraphics[width=14cm]{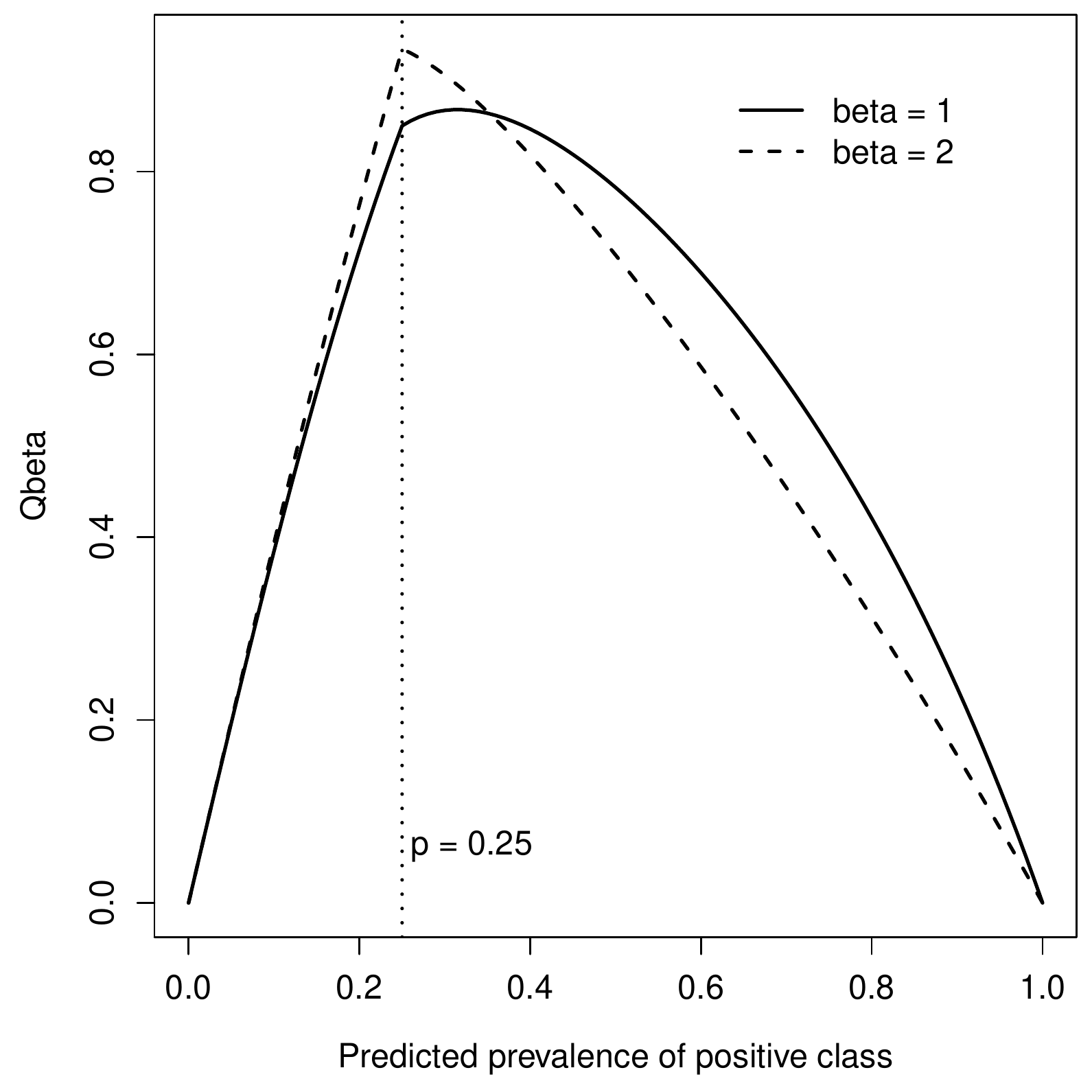}}
\fi
\end{center}
\end{figure}

Using the observation from Remark~\ref{rm:comments}~(iii), in Section~\ref{se:binormal} 
we demonstrate by the standard example of
a binormal model that optimal classifiers with respect to $Q_\beta$ in general are not best quantifiers.
First note that Remark~\ref{rm:comments}~(iii) implies the following result.
\begin{proposition}\label{pr:alter}
Under Assumption~\ref{as:0}, define $Q_\beta(H)$, $H \in \mathcal{H}$ by \eqref{eq:Qbeta} and 
$H_q$, $q\in (0,1)$ by \eqref{eq:shape}. For $0< u < 1$ denote by $q(u)$ the $u$-quantile\footnote{%
For a real random variable $X$ and $\alpha\in(0,1)$, the $\alpha$-quantile $q_\alpha(X)$ of $X$ is defined by
$q_\alpha(X)=\inf\{x\in\mathbb{R}: \mathrm{P}[X\le x]\ge \alpha\}$.\label{fn:quant}} of
$\mathrm{P}[A\,|\,\mathcal{H}]$. 
If the distribution of $\mathrm{P}[A\,|\,\mathcal{H}]$ is continuous then
it holds that
\begin{equation}\label{eq:sup}
 \underset{H\in\mathcal{H}}{\sup} Q_\beta(H) \ = \ \underset{\mathrm{P}[A]\le u < 1}{\sup} Q_\beta(H_{q(1-u)}).
\end{equation}
\end{proposition}
\textbf{Proof.} Observe that $\underset{H\in\mathcal{H}}{\sup} Q_\beta(H)
= \underset{H\in\mathcal{H},\, 0 < \mathrm{P}[H] < 1}{\sup} Q_\beta(H)$ since $Q_\beta(H) = 0$ 
for $\mathrm{P}[H]=0$ and $\mathrm{P}[H]=1$.

Fix $H\in\mathcal{H}$ with $0<\mathrm{P}[H] < 1$. 
By continuity of $x \mapsto \mathrm{P}\bigl[\mathrm{P}[A\,|\,\mathcal{H}] \le x\bigr]$, then
we have $\mathrm{P}[H_{q(u)}] = \mathrm{P}[H]$ for $u = 1-\mathrm{P}[H]$.
Hence 
Remark~\ref{rm:comments}~(iii) implies
\begin{align*}
 Q_\beta(H) & = \frac{1+\beta^2}
    {\frac{\beta^2}{\mathrm{NAS}(H_{q(u)})} + \frac1{\mathrm{P}[H\,|\,A]}}\\
    & \le \frac{1+\beta^2}
    {\frac{\beta^2}{\mathrm{NAS}(H_{q(u)})} + \frac1{\mathrm{P}[H_{q(u)}\,|\,A]}}\\
    & = Q_\beta(H_{q(u)}).
\end{align*}
This implies
\begin{equation*}
 \underset{H\in\mathcal{H}}{\sup} Q_\beta(H) \ = \ \underset{0 < u < 1}{\sup} Q_\beta(H_{q(1-u)}).
\end{equation*} 
However, for $0 < u \le \mathrm{P}[A]$ both $\mathrm{P}[H_{q(1-u)}\,|\,A]$ and $\mathrm{NAS}(H_{q(1-u)})$
are non-decreasing in $u$. Hence it follows that $Q_\beta(H_{q(1-u)}) \le Q_\beta(H_{q(1-\mathrm{P}[A])})$ for
$0 < u \le \mathrm{P}[A]$ and therefore \eqref{eq:sup}.
\hfill \qed

See Figure~\ref{fig:Qbeta} for an illustration of Proposition~\ref{pr:alter}. Unfortunately, in practice 
most of the time it is not possible to accurately estimate general posterior probabilities like
$\mathrm{P}[A\,|\,\mathcal{H}]$. This has led some authors to propose workarounds  like the one
by \citet{Platt99probabilisticoutputs} which do not necessarily deliver good results. However, the following
corollary describes a special setting in which the right-hand side of \eqref{eq:sup} can be
considerably simplified.

\begin{corollary}\label{co:alter}
Under Assumption~\ref{as:0}, define $Q_\beta(H)$, $H \in \mathcal{H}$ by \eqref{eq:Qbeta} and 
$H_q$, $q\in \mathbb{R}$ by \eqref{eq:shape}. Make these two additional assumptions:
\begin{itemize} 
\item[(i)] There is a real random variable $X$ on
$(\Omega, \mathcal{A}, \mathrm{P})$ with
continuous distribution such that $\mathcal{H} \supset\sigma(X)$, i.e.\ $X$ is $\mathcal{H}$-measurable.
\item[(ii)] There is a continuous function $f: \mathbb{R} \to [0,1]$  such that
\begin{equation*}
\mathrm{P}[A\,|\,\mathcal{H}] \ = \ f(X).
\end{equation*}
The function $f$ is either strictly increasing with\/
$\lim_{x\to -\infty} f(x) = 0$ and\/ $\lim_{x\to \infty} f(x) = 1$ or strictly decreasing with\/
$\lim_{x\to -\infty} f(x) = 1$ and\/ $\lim_{x\to \infty} f(x) = 0$. 
\end{itemize}
Then if $f$ is increasing it holds that
\begin{align*}
\underset{H\in\mathcal{H}}{\sup} Q_\beta(H) & \ = \ \underset{\mathrm{P}[A] \le u \le 1}{\sup} 
Q_\beta\bigl(\{X > q_{1-u}(X)\}\bigr).
\intertext{Otherwise, if $f$ is decreasing it holds that}
\underset{H\in\mathcal{H}}{\sup} Q_\beta(H) & \ = \ \underset{\mathrm{P}[A] \le u \le 1}{\sup} 
Q_\beta\bigl(\{X < q_{u}(X)\}\bigr).
\end{align*}
\end{corollary}
\textbf{Proof.} Follows from Proposition~\ref{pr:alter} because $H_{q(1-u)} = \{X > q_{1-u}(X)\}$ in the
case of increasing $f$ and $H_{q(1-u)} = \{X < q_{u}(X)\}$ in the case of decreasing $f$, for
$\mathrm{P}[A] \le u < 1$. \hfill \qed

Corollary~\ref{co:alter} allows us in Section~\ref{se:binormal} to replicate the experiment by
\citet{barranquero2015quantification} in a fully controlled environment such that the merits and
limitations of their approach can carefully be studied.

 
\section{The binormal case with equal variances}
\label{se:binormal}

We consider the `binormal model' with equal variances as 
an example that fits into the setting of Assumption~\ref{as:0} and Corollary~\ref{co:alter}.
\begin{itemize}
\item $\Omega = \mathbb{R} \times \{-1, 1\}$, $\mathcal{A} = \mathcal{B}(\mathbb{R})
    \otimes \mathcal{P}(\{-1, 1\})$ where $\mathcal{B}(\mathbb{R})$ denotes the Borel-$\sigma$-field on $\mathbb{R}$
    and $\mathcal{P}(\{-1, 1\})$ is the power set of $\{-1, 1\}$.
\item On $\Omega$, we define the projections $X$ and $Y$, i.e.\ for $\omega = (x, y) \in \Omega$ we let
    $X(\omega) = x$ and $Y(\omega) = y$.
\item $A = \{Y=1\} \notin \mathcal{H} = \sigma(X)$.
\item $\mathrm{P}$ is defined by specifying the marginal distribution of $Y$ with $\mathrm{P}[A] = p \in (0,1)$, 
     and defining the conditional distribution of $X$ given $Y$ as normal distributions 
     with equal variances:
\begin{subequations} 
\begin{equation}\label{eq:CondNormal}
\begin{split}
    \mathrm{P}[X \in \cdot\,|\,A] & = \mathcal{N}(\nu, \sigma^2),\\
    \mathrm{P}[X \in \cdot\,|\,A^c] & = \mathcal{N}(\mu, \sigma^2). 
\end{split}
\end{equation}
    In \eqref{eq:CondNormal}, we assume that $\mu < \nu$ and $\sigma > 0$. \eqref{eq:CondNormal} implies that the
    distribution of $X$ is given by a mixture of normal distributions\footnote{%
    $\Phi$ denotes the standard normal distribution function $\Phi(x) = \frac{1}{\sqrt{2\,\pi}} \int_{-\infty}^x 
    e^{-\,y^2/2}\,d y$.}
\begin{equation}
\mathrm{P}[X \le x] \ =\ p\,\Phi\left(\frac{x-\nu}{\sigma}\right) +
    (1-p)\,\Phi\left(\frac{x-\mu}{\sigma}\right), \quad x \in \mathbb{R}.
\end{equation}
\end{subequations}
\item The posterior probability $\mathrm{P}[A\,|\,\mathcal{H}]$ in this setting is given by
\begin{equation}
    \mathrm{P}[A\,|\,\mathcal{H}] \  = \ \frac{1}{1 + \exp(a\,X + b)},
\end{equation}
with $a = \frac{\mu-\nu}{\sigma^2} < 0$ and $b = \frac{\nu^2-\mu^2}{2\,\sigma^2} + \log\left(\frac{1-p}{p}\right)$.
\end{itemize}

We replicate the experiment of \citet{barranquero2015quantification} in this setting:
\begin{itemize}
\item We look at a `training set' $(\Omega, \mathcal{A}, \mathrm{P})$ and a `target dataset'
$(\Omega, \mathcal{A}, \mathrm{P}_1)$. The probability measures $\mathrm{P}$ and $\mathrm{P}_1$ are defined like
$\mathrm{P}$ above but with (possibly) different values of $\mathrm{P}[A] = p$ and $\mathrm{P}_1[A] = p_1$
respectively.
\item We ``train'' a classifier on the training set $(\Omega, \mathcal{A}, \mathrm{P})$ by maximising 
$Q_\beta$ as given by \eqref{eq:Qbeta}. 
Classifiers can be identified with sets $H \in \mathcal{H}$ in the sense that
the prediction is the positive class $A$ if $H$ occurs and the negative class $A^c$ otherwise. 
``Training'' in the binormal setting actually means to 
make use of Corollary~\ref{co:alter} in order to identify the optimal classifier $H^\ast$. 
The following formulae are used for the optimization of $Q_\beta$:
\begin{subequations}
\begin{align}
\mathrm{P}[X > q_{1-u}(X)\,|\,A] & = 1-\Phi\left(\frac{q_{1-u}(X)-\nu}{\sigma}\right),\\
\mathrm{NAS}^\ast(\{X > q_{1-u}(X)\}) & = 1 - \frac{\max(u-p,0)}{1-p} - \frac{\max(p-u,0)}{p}.
\label{eq:NAS.star2}
\end{align}
\end{subequations}
The quantile $q_{1-u}(X)$ (see footnote~\ref{fn:quant} for its definition) 
must be numerically determined by solving for $x$ the following equation:
\begin{equation}
1-u \ = \ p_0\,\Phi\left(\frac{x-\nu}{\sigma}\right) +
    (1-p_0)\,\Phi\left(\frac{x-\mu}{\sigma}\right).
\end{equation}
As the optimization problem has no closed-form solution, we apply the R-function 'optimize' \citep{RSoftware}
to find the optimal classifier $H^\ast = \{X > q_{1-u^\ast}(X)\}$.
\item We evaluate the $Q_\beta$-optimal classifier $H^\ast$ on the target dataset $(\Omega, \mathcal{A}, \mathrm{P}_1)$ by
calculating $\mathrm{P}_1[H^\ast]$ and compare its value to $p_1 = \mathrm{P}_1[A]$ in order to check how 
good the Classify \& Count approach \citep{forman2008quantifying} based on $H^\ast$ is. We compare
the performance of $H^\ast$ with two other classifiers: the minimax classifier $H_{\mathrm{Mini}}$ from
Corollary~\ref{co:best} and the locally best classifier $H_{\mathrm{loc}}$ from Corollary~\ref{co:localbest}.
This means that we need to evaluate \eqref{eq:error} for the three classifiers.
For this purpose, the following formulae are used:
\begin{subequations}
\begin{align}
    \mathrm{P}[H^\ast\,|\,A] & = \Phi\left(\frac{q_{1-u^\ast}(X)-\nu}{\sigma}\right), & 
    \mathrm{P}[H^\ast\,|\,A^c] & =\Phi\left(\frac{q_{1-u^\ast}(X)-\mu}{\sigma}\right);\\
    \mathrm{P}[H_{\mathrm{Mini}}\,|\,A] & = \Phi\left(\frac{\frac{\mu+\nu}2-\nu}{\sigma}\right), & 
    \mathrm{P}[H_{\mathrm{Mini}}\,|\,A^c] & =\Phi\left(\frac{\frac{\mu+\nu}2-\mu}{\sigma}\right);\\
    \mathrm{P}[H_{\mathrm{loc}}\,|\,A] & = \Phi\left(\frac{q_{1-p}(X)-\nu}{\sigma}\right), & 
    \mathrm{P}[H_{\mathrm{loc}}\,|\,A^c] & =\Phi\left(\frac{q_{1-p}(X)-\mu}{\sigma}\right).
\end{align}
\end{subequations}
\item For the calculations, we have used the following parameters: 
\begin{equation}\label{eq:par}
\mu=0,\quad \nu=2,\quad \sigma=1,\quad
\mathrm{P}[A] = p = 25\%.
\end{equation}
\end{itemize}
The results of the calculations are shown in Figures~\ref{fig:Qbeta} and \ref{fig:PredictionError}. Figure~\ref{fig:Qbeta}
presents graphs of $u \mapsto Q_\beta\bigl(\{X > q_{1-u}(X)\}\bigr)$ in the binormal setting of this section, with
parameters chosen as in \eqref{eq:par}. The Q-measure $Q_\beta$ is defined by \eqref{eq:Qalter}, but we use 
$\mathrm{NAS}^\ast$ instead of $\mathrm{NAS}$.
The solid curve is for $\beta =1$, where equal weights are put on $\mathrm{NAS}^\ast$ and the TPR $\mathrm{P}[H\,|\,A]$,
and the dashed curve is for $\beta =2$, where the weight put on $\mathrm{NAS}^\ast$ is four times the weight for the TPR.

The kinks in both graphs of Figure~\ref{fig:Qbeta} are due to the fact that the mapping
$u \mapsto \mathrm{NAS}^\ast(\{X > q_{1-u}(X)\})$
with $\mathrm{NAS}^\ast(\{X > q_{1-u}(X)\})$ defined by the right-hand side of \eqref{eq:NAS.star2} 
is not differentiable in $u = p$. The function $u \mapsto \mathrm{NAS}$ would not be differentiable in $u = p$ either.
Both curves have a unique maximum which is at $u = p = 0.25$ for $\beta =2$ and at some $u > p=0.25$ for $\beta = 1$.

Hence the graph for $\beta =2$ has its maximum at that $u$ where $\mathrm{NAS}^\ast(\{X > q_{1-u}(X)\})$ takes its maximum 
value 1. As a consequence, in the case $\beta =2$, the $Q_\beta$-optimal classifier is identical with the locally 
best classifier according to Corollary~\ref{co:localbest}.

In contrast, in the case $\beta=1$, for $u$ slightly greater than $p = 0.25$ the decline in value of 
$\mathrm{NAS}^\ast(\{X > q_{1-u}(X)\})$ is over-compensated by a rise in the value of the TPR such
that the maximum is incurred at some $u > p$. The consequence of this
for the error \eqref{eq:error} in the prediction of a target dataset positive class prevalence is
displayed in Figure~\ref{fig:PredictionError}.

Figure~\ref{fig:PredictionError} shows that by incident the error-performance of the $Q_\beta$-optimal classifier for
$\beta = 1$ is close to the performance of the minimax classifier identified in Corollary~\ref{co:best}.
Nonetheless, its performance would be deemed unsatisfactory if the true positive class prevalence in the target dataset
were the same or nearly the same as the assumed positive class prevalence of 25\% in the training dataset.
In that case, clearly the locally best classifier as identified by the $Q_\beta$-measure for $\beta =2$ or by
Corollary~\ref{co:localbest} would perform much better and even perfectly if the training and target 
prevalences were the same.


\section{Conclusions}
\label{se:concl}

We have investigated a claim by \citet{barranquero2015quantification} and \citet{esuli2015optimizing} that
binary class prevalences on target datasets can be estimated by classifiers without further adjustments if
these classifiers are developed as so-called quantifiers on the training datasets. The development
of such quantifiers involves special optimisation criteria covering both calibration (i.e.\ the number
of objects predicted as positive equals the true number of positives) and classification power.

\citet{barranquero2015quantification} recommended the so-called Q-measure as the optimisation criterion
for the quantifier and tested their approach on some real-world datasets. It is not fully clear, however,
from \citet{barranquero2015quantification} which of their observations are fundamental and which are
sample-driven and incidental. In this paper, therefore, we have identified the theoretically correct way
to determine the best quantifiers according to the Q-measure criterion. We then have replicated the
experiment of \citet{barranquero2015quantification} in the fully controlled setting of the binormal model
with equal variances. For binary classification settings, we have found that 
\begin{itemize}
\item[1)] quantification without adjustments in principle may work if proper calibration on 
the training dataset is ensured and the positive class prevalences 
on the training and target datasets are the same or nearly the same, and
\item[2)]  
 the Q-measure approach has to be deployed with care 
because the quantifiers resulting from it may be miscalibrated.
\end{itemize}
All in all, 
these findings suggest that the potential applications of quantification without 
adjustments are rather limited.



\addcontentsline{toc}{section}{References}

\appendix

\section{Appendix: Optimal classifiers for the F-measure}
\label{se:OptClass}

In the following, we give a proof of \eqref{eq:Fgen} which is based on the
idea of \citet{ICML2012Ye_175} and fills the gaps left by them.

\textbf{Proof of \eqref{eq:Fgen}.} The proof makes use of the four following lemmata.

\begin{lemma}\label{le:ext}
Let $(\Omega, \mathcal{A}, \mathrm{P})$, $A \in \mathcal{A}$ and $\mathcal{H} \subset \mathcal{A}$ be 
as in Assumption~\ref{as:0}.Then there are a probability space $(\tilde{\Omega}, \tilde{\mathcal{A}}, 
\tilde{\mathrm{P}})$, an event $\tilde{A} \in \tilde{\mathcal{A}}$ and a sub-$\sigma$-field $\tilde{\mathcal{H}}$ of
$\tilde{\mathcal{A}}$ with the following three properties:
\begin{itemize}
\item[(i)] There is a measurable mapping $X: (\tilde{\Omega}, \tilde{\mathcal{A}}) \to (\Omega, \mathcal{A})$
such that 
$$\tilde{\mathrm{P}}\circ X^{-1} = \mathrm{P},\quad \tilde{A} = X^{-1}(A),\quad
\tilde{\mathcal{H}} \supset X^{-1}(\mathcal{H}), \quad \tilde{\mathrm{P}}[\tilde{A}\,|\,\tilde{\mathcal{H}}] =
\mathrm{P}[A\,|\,\mathcal{H}]\circ X.$$
\item[(ii)] $(\tilde{\Omega}, \tilde{\mathcal{A}}, 
\tilde{\mathrm{P}})$, $\tilde{A} \in \tilde{\mathcal{A}}$ and $\tilde{\mathcal{H}}$ satisfy Assumption~\ref{as:0}.
\item[(iii)] For any $H\in \mathcal{H}$ and $\alpha \in [0,1]$
there exists $\tilde{H}_\alpha \in \tilde{\mathcal{H}}$, $\tilde{H}_\alpha \subset X^{-1}(H)
    \stackrel{\mathrm{def}}{=} \tilde{H}$ such that
$$\tilde{\mathrm{P}}[\tilde{H}_\alpha] = \alpha\,\tilde{\mathrm{P}}[\tilde{H}].$$
\end{itemize}
\end{lemma}
\textbf{Proof of Lemma~\ref{le:ext}.} For (i) and (ii), define $\tilde{\Omega} = \Omega \times (0,1)$, 
$\tilde{\mathcal{A}} = \mathcal{A} \otimes \mathcal{B}(0,1)$, $\tilde{\mathrm{P}} = \mathrm{P} \otimes U(0,1)$, 
and let $X$ be the projection from  $\tilde{\Omega}$ to $\Omega$, where $\mathcal{B}(0,1)$ denotes the 
Borel-$\sigma$-field on $(0,1)$ and $U(0,1)$ stands for the uniform distribution on $(0,1)$. 
In addition, let $\tilde{\mathcal{H}} = \mathcal{H} \otimes \mathcal{B}(0,1)$.
For (iii), define $\tilde{H}_\alpha = H \times (0,\alpha)$. \hfill \qed

\begin{lemma}\label{le:quant}
Under Assumption~\ref{as:0}, for any fixed $H\in\mathcal{H}$ 
there is a number $q\in[0,1]$ such that
\begin{equation}\label{eq:ineq}
    \mathrm{P}\bigl[\mathrm{P}[A\,|\,\mathcal{H}] > q\bigr] \le \mathrm{P}[H] \le 
\mathrm{P}\bigl[\mathrm{P}[A\,|\,\mathcal{H}] \ge q\bigr].
\end{equation}
\end{lemma}
\textbf{Proof.} Choose $q$ as an $(1-\mathrm{P}[H])$-quantile of $\mathrm{P}[A\,|\,\mathcal{H}]$ or $q \in \{0,1\}$.
\hfill \qed

\begin{lemma}\label{le:leq}
Under Assumption~\ref{as:0}, if for a fixed $H\in\mathcal{H}$ there are $H^\ast\in\mathcal{H}$ and $0\le q\le 1$
such that 
$$\{\mathrm{P}[A\,|\,\mathcal{H}] > q\} \subset H^\ast \subset \{\mathrm{P}[A\,|\,\mathcal{H}] \ge q\}\quad
\text{and}\quad \mathrm{P}[H] = \mathrm{P}[H^\ast],$$
then it holds that $\mathrm{P}[H \cap A] \le \mathrm{P}[H^\ast \cap A]$.
\end{lemma}
\textbf{Proof.} By assumption we have $H^\ast \subset \{\mathrm{P}[A\,|\,\mathcal{H}] \ge q\}$ and
$(H^\ast)^c \subset \{\mathrm{P}[A\,|\,\mathcal{H}] \le q\}$. This implies
\begin{align*}
\mathrm{P}[H^\ast \cap A] - \mathrm{P}[H \cap A] & = 
    \mathrm{P}[H^\ast \cap H^c \cap A] - \mathrm{P}[(H^\ast)^c \cap H \cap A]\\
    & = \mathrm{E}\bigl[\mathrm{P}[A\,|\,\mathcal{H}]\,\mathbf{1}_{H^\ast \cap H^c}\bigr] -
        \mathrm{E}\bigl[\mathrm{P}[A\,|\,\mathcal{H}]\,\mathbf{1}_{(H^\ast)^c \cap H}\bigr]\\
    & \ge q\,\mathrm{P}[H^\ast \cap H^c] - q\,\mathrm{P}[(H^\ast)^c \cap H]\\
    & = q\,\bigl(\mathrm{P}[H^\ast]-\mathrm{P}[H]\bigr) = 0.\hspace{5cm} \Box
\end{align*}

\begin{lemma}\label{le:main}
Under Assumption~\ref{as:0}, for all $\beta > 0$ and $q \in [0,1]$ it holds that
\begin{equation}\label{eq:step}
    F_\beta(H)\  \le  \ 
 \max\bigl(F_\beta(H\cap \{\mathrm{P}[A\,|\,\mathcal{H}] > q\}), 
	F_\beta(H \cup \{\mathrm{P}[A\,|\,\mathcal{H}] = q\})\bigr).
\end{equation}
\end{lemma}
\textbf{Proof.} The proof of this lemma is essentially identical with the proof given by 
\citet{ICML2012Ye_175} for Theorem~4. This suggests that the proof by \citet{ICML2012Ye_175} is incomplete 
since some more steps are needed to show that \eqref{eq:step} implies \eqref{eq:Fgen}.

Fix $\beta > 0$. We distinguish the two cases $q = 1$ and $0 \le q < 1$.
For the sake of a more concise notation, we define 
$$Z = \mathrm{P}[A\,|\,\mathcal{H}].$$

\textbf{Case $\mathbf{q = 1}$.} We show that $F_\beta(H) \le F_\beta(H \cup \{Z = 1\})$.
Note that $H \cup \{Z = 1\} = H \cup (H^c \cap \{Z = 1\})$ and that 
$$\mathrm{P}[A\cap H^c \cap \{Z = 1\}] = \mathrm{E}\bigl[Z\, 
    \mathbf{1}_{H^c\cap\{Z= 1\}}\bigr] = \mathrm{P}[H^c \cap \{Z= 1\}].$$
This implies
\begin{align*}
F_\beta(H) & \le F_\beta(H \cup \{Z = 1\})\\
\Leftrightarrow\ \frac{(1+\beta^2)\,\mathrm{P}[A\cap H]}{\beta^2\,\mathrm{P}[A] + \mathrm{P}[H]} & \le 
\frac{(1+\beta^2)\,\bigl(\mathrm{P}[A\cap H] + \mathrm{P}[H^c \cap \{Z = 1\}]\bigr)}
{\beta^2\,\mathrm{P}[A] + \mathrm{P}[H] + \mathrm{P}[H^c \cap \{Z = 1\}]}\\
\Leftrightarrow\ \mathrm{P}[A\cap H]\,\mathrm{P}[H^c \cap \{Z = 1\}] & \le 
\mathrm{P}[H]\,\mathrm{P}[H^c \cap \{Z = 1\}] + \beta^2 \, \mathrm{P}[A]\,\mathrm{P}[H^c \cap \{Z = 1\}].
\end{align*}
The last row of this equation chain is true because $\mathrm{P}[A\cap H] \le \mathrm{P}[H]$. This proves
\eqref{eq:step} in the case $q = 1$.

\textbf{Case $\mathbf{0 \le q < 1}$.} We show that $F_\beta(H) >  F_\beta(H\cap \{Z > q\})$
implies $F_\beta(H) \le 
	F_\beta(H \cup \{Z = q\})$. Observe first that 
\begin{gather}
F_\beta(H)\ >\  F_\beta(H\cap \{Z > q\}) \notag \\
\Leftrightarrow \ \bigl(\beta^2 \,\mathrm{P}[A] + \mathrm{P}\bigl[H\cap \{Z > q\}\bigr]\bigr)\, 
    \mathrm{P}\bigl[A\cap H\cap \{Z \le q\}\bigr] \ >\ 
\mathrm{P}\bigl[H\cap \{Z \le q\}\bigr]\,\mathrm{P}\bigl[A\cap H\cap \{Z > q\}\bigr]. \label{eq:equiv}  
\end{gather}
Secondly, we have
\begin{gather}
F_\beta(H) \ \le\ 	F_\beta(H \cup \{Z = q\})\notag\\
\Leftrightarrow \ 
\mathrm{P}[A\cap H] \, \mathrm{P}\bigl[H^c\cap \{Z = q\}\bigr] \ \le \
\bigl(\beta^2 \,\mathrm{P}[A] + \mathrm{P}[H]\bigr)\,q\,\mathrm{P}\bigl[H^c\cap \{Z = q\}\bigr]. 
\label{eq:second}
\end{gather}
If $\mathrm{P}\bigl[H^c\cap \{Z = q\}\bigr] = 0$ then by \eqref{eq:second} there is nothing left to be proved.
Hence assume $\mathrm{P}\bigl[H^c\cap \{Z = q\}\bigr] > 0$. Then from \eqref{eq:second} we obtain
\begin{gather}
F_\beta(H) \ \le\ 	F_\beta(H \cup \{Z = q\})\notag\\
\Leftrightarrow \ 
\mathrm{P}[A\cap H] \ \le \
\bigl(\beta^2 \,\mathrm{P}[A] + \mathrm{P}[H]\bigr)\,q. 
\label{eq:next}
\end{gather}
\eqref{eq:equiv} implies $\mathrm{P}\bigl[H\cap \{Z \le q\}\bigr] > 0$ because otherwise we would have $0 > 0$.
Therefore, \eqref{eq:next} follows from \eqref{eq:equiv} because
\begin{align*}
\mathrm{P}[A\cap H] & = \mathrm{P}\bigl[A\cap H\cap \{Z \le q\}\bigr] + \mathrm{P}\bigl[A\cap H\cap \{Z > q\}\bigr]\\
& \le \mathrm{P}\bigl[A\cap H\cap \{Z \le q\}\bigr] + 
    \frac{\bigl(\beta^2 \,\mathrm{P}[A] + \mathrm{P}\bigl[H\cap \{Z > q\}\bigr]\bigr)\, 
    \mathrm{P}\bigl[A\cap H\cap \{Z \le q\}\bigr]}{\mathrm{P}\bigl[H\cap \{Z \le q\}\bigr]}\\
& = \mathrm{E}\bigl[Z\, 
    \mathbf{1}_{H\cap\{Z\le q\}}\bigr] \left(1 + \frac{\beta^2 \,\mathrm{P}[A] + 
    \mathrm{P}\bigl[H\cap \{Z > q\}\bigr]}{\mathrm{P}\bigl[H\cap \{Z \le q\}\bigr]}\right)\\
& \le \mathrm{P}\bigl[H\cap \{Z \le q\}\bigr]\,q\, 
    \frac{\beta^2 \,\mathrm{P}[A] + \mathrm{P}[H]}{\mathrm{P}\bigl[H\cap \{Z \le q\}\bigr]}\\
& = \bigl(\beta^2 \,\mathrm{P}[A] + \mathrm{P}[H]\bigr)\,q.
\end{align*}
Hence  \eqref{eq:next} is true if \eqref{eq:equiv} holds. This completes the proof of \eqref{eq:step}. \hfill\qed

\textbf{Finishing the proof of \eqref{eq:Fgen}.} Fix $H \in \mathcal{H}$. We need to show that there is a number $0 \le q \le 1$ such
that 
\begin{equation}\label{eq:Faux}
F_\beta(H)  \ \le \ 
	\max\bigl(F_\beta(\{\mathrm{P}[A\,|\,\mathcal{H}] > q\}), 
	F_\beta(\{\mathrm{P}[A\,|\,\mathcal{H}] \ge q\})\bigr).
\end{equation}
Since $F_\beta(H) = 0$ if $\mathrm{P}[H] = 0$, \eqref{eq:Faux} is obvious is that case. Therefore, we may assume
$\mathrm{P}[H] > 0$ for the remainder of the proof.

The notation $F_\beta(H)$ hides the fact that $F_\beta(H)$ does not only depend on the classifier $H$, but also on the 
class event $A$ and the probability measure $\mathrm{P}$. This matters because the next step in the proof is 
to replace $(\Omega, \mathcal{A}, \mathrm{P})$, $A$ and $\mathcal{H}$ by $(\tilde{\Omega}, \tilde{\mathcal{A}}, 
\tilde{\mathrm{P}})$, $\tilde{A}$ and $\tilde{\mathcal{H}}$ as provided by Lemma~\ref{le:ext}. In the following, 
when using notation like $F_\beta(\tilde{H})$ we implicitly assume that all ingredients for the calculation come 
from the probability space $(\Omega, \mathcal{A}, \mathrm{P})$.

Against this backdrop, define $\tilde{H} = X^{-1}(H)$ where $X$ denotes the projection on $\Omega$ as defined in 
Lemma~\ref{le:ext}. Then it is easy to see that
$$F_\beta(H) =  F_\beta(\tilde{H}), \ F_\beta(\{\mathrm{P}[A\,|\,\mathcal{H}] > q\}) =
    F_\beta(\{\tilde{\mathrm{P}}[\tilde{A}\,|\,\tilde{\mathcal{H}}] > q\}), \ 
    F_\beta(\{\mathrm{P}[A\,|\,\mathcal{H}] \ge q\}) =
    F_\beta(\{\tilde{\mathrm{P}}[\tilde{A}\,|\,\tilde{\mathcal{H}}] \ge q\}).$$
Hence, if we prove
\begin{equation}\label{eq:Fmod}
F_\beta(\tilde{H})  \ \le \ 
	\max\bigl(F_\beta(\{\tilde{\mathrm{P}}[\tilde{A}\,|\,\tilde{\mathcal{H}}] > q\}), 
	F_\beta(\{\tilde{\mathrm{P}}[\tilde{A}\,|\,\tilde{\mathcal{H}}] \ge q\})\bigr),
\end{equation}
\eqref{eq:Faux} immediately follows. Choose $0\le q \le 1$ according to Lemma~\ref{le:quant} such that
\begin{equation}\label{eq:ineq.0}
    \tilde{\mathrm{P}}\bigl[\tilde{\mathrm{P}}[\tilde{A}\,|\,\tilde{\mathcal{H}}] > q\bigr] \le 
    \tilde{\mathrm{P}}[\tilde{H}] \le 
\tilde{\mathrm{P}}\bigl[\tilde{\mathrm{P}}[\tilde{A}\,|\,\tilde{\mathcal{H}}] \ge q\bigr].
\end{equation}
By Lemma~\ref{le:ext}~(iii) and \eqref{eq:ineq.0}, 
there exists an event $\tilde{H}_0 \in \tilde{\mathcal{H}}$ such that
$$\tilde{H}_0 \ \subset \ \{\tilde{\mathrm{P}}[\tilde{A}\,|\,\tilde{\mathcal{H}}] = q\}\quad
    \text{and} \quad \tilde{\mathrm{P}}[\tilde{H}_0] \ = \ \tilde{\mathrm{P}}[\tilde{H}] -
    \tilde{\mathrm{P}}\bigl[\tilde{\mathrm{P}}[\tilde{A}\,|\,\tilde{\mathcal{H}}] > q\bigr].$$
Hence for $\tilde{H}^\ast = \tilde{H}_0 \cup \{\tilde{\mathrm{P}}[\tilde{A}\,|\,\tilde{\mathcal{H}}] > q\}$ we
obtain
$$\tilde{\mathrm{P}}[\tilde{H}^\ast] \ = \ \tilde{\mathrm{P}}[\tilde{H}] 
    \quad\text{and} \quad \{\tilde{\mathrm{P}}[\tilde{A}\,|\,\tilde{\mathcal{H}}] > q\} \ \subset \
    \tilde{H}^\ast \ \subset \ \{\tilde{\mathrm{P}}[\tilde{A}\,|\,\tilde{\mathcal{H}}] \ge q\}.$$
Now, Lemma~\ref{le:leq} implies that
\begin{equation}\label{eq:interim}
F_\beta(\tilde{H})\ =\ \frac{(1+\beta^2)\,\tilde{\mathrm{P}}[\tilde{A}\cap\tilde{H}]}
    {\beta^2\,\tilde{\mathrm{P}}[\tilde{A}] + \tilde{\mathrm{P}}[\tilde{H}]}\ \le\
    \frac{(1+\beta^2)\,\tilde{\mathrm{P}}[\tilde{A}\cap\tilde{H}^\ast]}
    {\beta^2\,\tilde{\mathrm{P}}[\tilde{A}] + \tilde{\mathrm{P}}[\tilde{H}^\ast]}\ =\
    F_\beta(\tilde{H}^\ast).
\end{equation}
From Lemma~\ref{le:main}, it follows that
\begin{equation}\label{eq:step.0}
    F_\beta(\tilde{H}^\ast)\  \le  \ 
 \max\bigl(F_\beta(\tilde{H}^\ast\cap \{\tilde{\mathrm{P}}[\tilde{A}\,|\,\tilde{\mathcal{H}}] > q\}), 
	F_\beta(\tilde{H}^\ast\cup \{\tilde{\mathrm{P}}[\tilde{A}\,|\,\tilde{\mathcal{H}}] = q\})\bigr).
\end{equation}
Note that by construction, we have got that
$$\tilde{H}^\ast \cap \{\tilde{\mathrm{P}}[\tilde{A}\,|\,\tilde{\mathcal{H}}] > q\} = 
    \{\tilde{\mathrm{P}}[\tilde{A}\,|\,\tilde{\mathcal{H}}] > q\} \quad \text{and}\quad
   \tilde{H}^\ast \cup \{\tilde{\mathrm{P}}[\tilde{A}\,|\,\tilde{\mathcal{H}}] = q\} =
   \{\tilde{\mathrm{P}}[\tilde{A}\,|\,\tilde{\mathcal{H}}] \ge q\}.$$
Hence,  \eqref{eq:interim} and \eqref{eq:step.0} together imply  \eqref{eq:Fmod}. \hfill \qed 
\end{document}